\documentclass[10pt,letterpaper,twocolumn]{article}

\usepackage[margin=1in]{geometry}
\usepackage[T1]{fontenc}
\usepackage[hyphens]{url}
\usepackage{graphicx}
\usepackage{xcolor}
\usepackage{amsmath,amssymb}
\usepackage{booktabs}
\usepackage{array}
\usepackage{multirow}
\usepackage{caption}
\usepackage{algorithm}
\usepackage{algorithmic}
\usepackage{natbib}
\usepackage{authblk}
\usepackage[hidelinks]{hyperref}
\usepackage{microtype}
\usepackage{newtxtext,newtxmath}

\title{\bfseries EvoPINN: Agentic Discovery of Executable Algorithms for Physics-Informed Neural Networks}

\author[1,2]{Peng Yin}
\author[2]{Kai Li\thanks{Corresponding author}}
\author[2]{Yifan Zhang}
\author[2]{Jian Cheng}

\affil[1]{School of Advanced Interdisciplinary Sciences, University of Chinese Academy of Sciences, Beijing, China}
\affil[2]{Institute of Automation, Chinese Academy of Sciences, Beijing, China}
\date{}

\begin{document}
\maketitle

\begin{abstract}
Physics-informed neural networks (PINNs) have emerged as a powerful paradigm for solving partial differential equations (PDEs), yet their performance heavily relies on the manual, trial-and-error engineering of neural representations, loss formulations, and optimization dynamics. While Large Language Models (LLMs) offer a promising avenue for automated design, unconstrained code generation often yields mathematically invalid or numerically unstable solutions under strict scientific computing constraints. To bridge this gap, we propose \textbf{EvoPINN}, an agentic framework that reformulates PINN development from labor-intensive manual design into a rigorous, execution-grounded algorithm discovery problem. EvoPINN navigates a modular search space by decoupling neural representations from training programs, utilizing an LLM agent to iteratively propose memory-conditioned programmatic modifications. To ensure scientific validity, all candidates undergo strict structural verification and budget-matched PDE evaluation. Extensive experiments across diverse PDE regimes (oscillatory, elliptic, dissipative, and nonlinear transport) demonstrate that EvoPINN discovers PDE-specialized learning algorithms that significantly reduce relative $L_{2}$ error compared to baselines. Crucially, EvoPINN autonomously invented SLRC-PINN, a novel architecture whose performance gains persist under rigorous parameter-matched comparisons, establishing the viability of execution-grounded agents for discovering genuinely new scientific computing mechanisms.
\end{abstract}

\section{Introduction}

Physics-informed neural networks (PINNs) provide a flexible, mesh-free framework for solving partial differential equations (PDEs) by incorporating physical laws into neural-network training \cite{raissi2019physics,lu2021deepxde}. Despite their conceptual elegance, their performance depends strongly on PDE-specific choices regarding coordinate representations, network architectures, loss formulations, collocation sampling, and optimization algorithms. Different PDE regimes often favor different mechanisms: oscillatory solutions benefit from frequency-aware representations, whereas nonlinear transport problems require stable optimization and enhanced resolution near steep gradients. Consequently, adapting PINNs to these distinct physical behaviors requires structural and programmatic interventions, making it a labor-intensive algorithm design problem with no single design dominating across diverse PDEs~\cite{wang2021gradient,wang2022ntk,hao2024pinnacle}.

Existing automated PINN methods primarily focus on tuning hyperparameters or assembling predefined components within human-specified search spaces~\cite{wang2023autopinn,wang2024naspinn, zhang2024discovering}. More recently, Large Language Model (LLM)-based systems (e.g., PINNsAgent~\cite{wuwu2025pinnsagent}, Lang-PINN~\cite{he2025langpinn}) have advanced this frontier by automating PINN pipeline construction and configuration selection via natural-language reasoning and experimental feedback. While these methods substantially reduce human effort, their exploration remains fundamentally constrained by human-specified templates and bounded design spaces. They excel at identifying optimal configurations among known alternatives, but do not readily modify the neural representation and training logic that jointly define the PINN learning algorithm. Consequently, the broader ambition—discovering genuinely new, executable learning mechanisms—remains largely unexplored for PINNs. Furthermore, directly leveraging LLMs for open-ended algorithmic generation in scientific computing presents a critical challenge: unconstrained LLMs frequently produce mathematically invalid, structurally redundant, or numerically unstable code when subjected to the strict physical and computational constraints of PDEs.

To bridge this gap, we introduce \textbf{EvoPINN}, an execution-grounded agentic framework for automatically discovering complete PINN learning algorithms. To tame the instability of open-ended LLM generation and solve the credit assignment problem during evolution, EvoPINN employs a modular design that decouples the neural representation module from the physics-informed training program. An LLM agent iteratively proposes memory-guided programmatic modifications to one module at a time. Crucially, EvoPINN ensures scientific validity by subjecting all proposals to a strict verification pipeline—including structural verification, executable testing, and budget-matched PDE evaluation—updating the algorithm only when a candidate exhibits structural novelty and reduces solution error. This controlled search process enables EvoPINN to explore representation and training mechanisms beyond a predefined library of PINN configurations.

We evaluate EvoPINN across oscillatory, elliptic, dissipative, and nonlinear-transport PDE regimes. The discovered algorithms consistently improve upon the baseline PINN and achieve performance comparable to or better than strong expert-designed methods across diverse problems. Furthermore, our evaluations confirm the robustness of the discovered mechanisms, demonstrating applicability to unseen equation parameters. Most notably, EvoPINN discovers SLRC-PINN for Burgers1D—a global–local architecture in which an adaptive bank of localized bases gates a zero-initialized corrective branch. To our knowledge, this specific topology has not been reported, and its advantage persists under parameter-matched comparisons with related PINN architectures. Together, these results demonstrate that execution-grounded agents can move automated PINN design beyond hyperparameter tuning toward discovering novel, effective, and reusable learning mechanisms.

\begin{figure*}[t]
\centering
\includegraphics[width=0.98\linewidth]{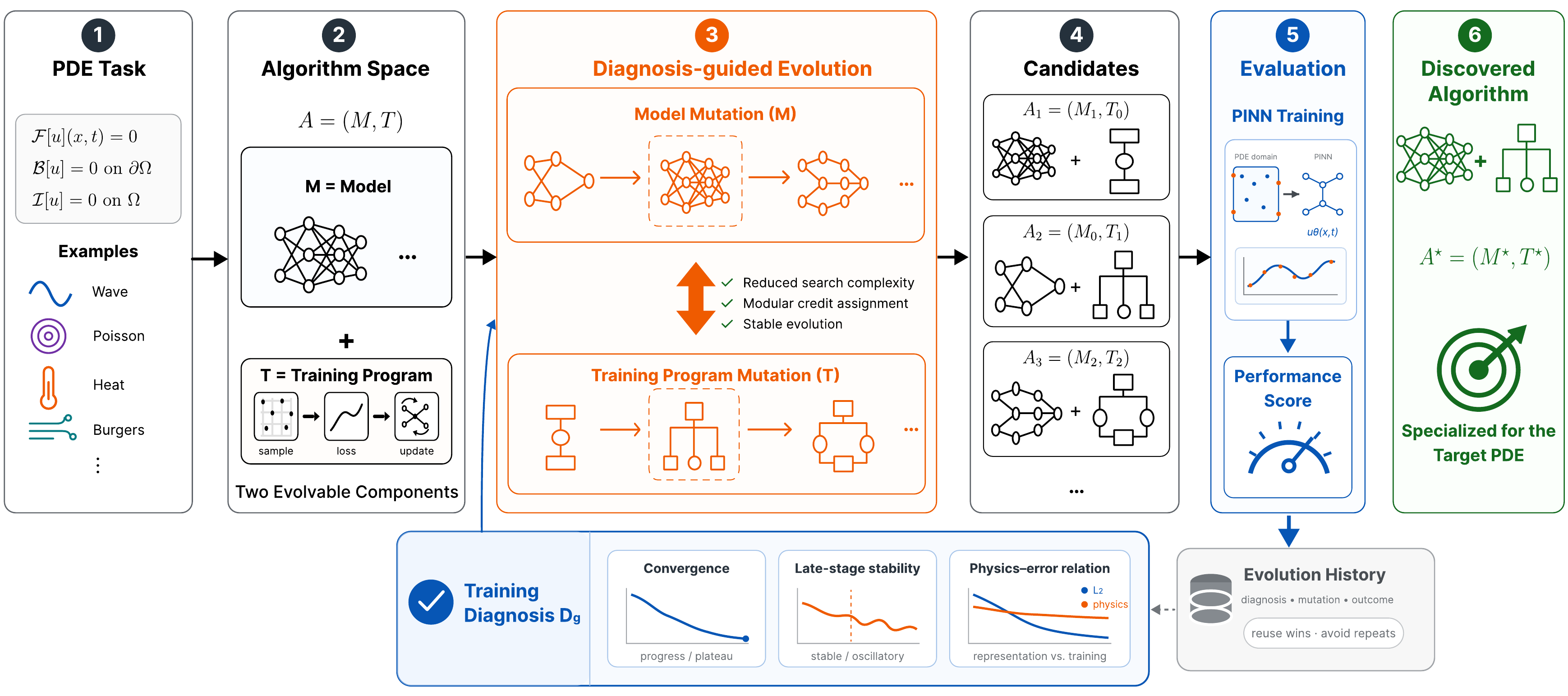}
\caption{Overview of EvoPINN. A PINN algorithm is represented as $A=(M,T)$, where $M$ defines the neural representation and $T$ defines the training program. At each generation, EvoPINN selects one module for modification, uses the current training behavior and earlier attempts to guide the LLM, and updates the current algorithm only when a generated program runs successfully under the shared training budget and reduces the solution error.}
\label{fig:overview}
\end{figure*}

\section{Related Work}

\subsection{Manual Mechanism Design in PINNs}

A large body of work improves Physics-Informed Neural Networks (PINNs) through manually designed representation topologies and training algorithms. Representation-oriented methods introduce adaptive activation functions \cite{jagtap2020adaptive}, frequency-aware feature encodings for high-frequency or multiscale dynamics \cite{sitzmann2020siren,wang2021eigenvector}, spatial feature embeddings \cite{tancik2020fourier}, and domain-decomposition architectures \cite{jagtap2020xpinn,moseley2023fbpinn}. Training-oriented methods focus on adaptive collocation sampling strategies \cite{daw2022rethinking,wu2023comprehensive}, residual objective modifications \cite{yu2022gpinn}, dynamic loss balancing \cite{wang2021gradient,wang2022ntk}, and multi-stage or curriculum optimization schemes \cite{krishnapriyan2021characterizing,yao2023multiadam}. Comprehensive benchmark evaluations demonstrate that no single manually designed PINN variant uniformly dominates across diverse physical regimes \cite{hao2024pinnacle}. Consequently, adapting PINNs to specific physical behaviors remains an arduous, trial-and-error engineering process, underscoring the critical imperative to transition from hand-crafted mechanisms to automated algorithmic discovery.

\subsection{Automated and Agentic PINN Design}

Earlier automated PINN design methods focus on optimizing discrete structural or numerical choices within predefined, highly restricted search spaces. Auto-PINN performs staged hyperparameter search over network depth, width, and optimizer configurations \cite{wang2023autopinn}, whereas NAS-PINN utilizes differentiable architecture search to select optimal connectivity patterns \cite{wang2024naspinn}. Beyond standard NAS, classical evolutionary computation approaches, such as DPSTE \cite{zhang2024discovering}, explore network structures alongside tree-based parametric activation functions using dynamic population scheduling. More recently, LLM-driven agentic frameworks extend automation to workflow and pipeline construction: PINNsAgent leverages empirical execution feedback to navigate architectures and hyperparameter configurations \cite{wuwu2025pinnsagent}, while Lang-PINN generates executable code pipelines directly from natural-language PDE specifications \cite{he2025langpinn}. However, these approaches remain constrained to selecting configurations, tuning hyperparameters, or composing workflows within pre-engineered templates. EvoPINN instead enables the open-ended discovery of novel PINN learning mechanisms.

\subsection{LLM-Guided Solver and Algorithm Discovery}

LLMs have increasingly been used to generate and refine scientific programs. In the PDE domain, CodePDE and PDE-SHARP synthesize solver implementations through reasoning or execution feedback, but focus on general PDE solvers rather than PINN-specific learning algorithms \cite{li2025codepde,fazliani2025pdesharp}. Beyond PDE solving, LLM-guided executable search has been applied to mathematical programs, combinatorial heuristics, reward functions, and optimization algorithms \cite{romera-paredes_mathematical_2024,liu2024eoh,ma2024eureka,ye2024reevo}. These methods typically evolve a bounded component, such as a heuristic, reward, or update function, within a fixed algorithmic scaffold.
EvoPINN breaks this limitation by elevating the search object to the complete PINN learning algorithm, which is significantly more challenging.

\section{Preliminaries of PINNs}

Consider a PDE instance $\mathcal P$ defined on a spatio-temporal domain $\Omega$, governed by a differential operator $\mathcal F[u]=0$ alongside boundary and initial constraints $\mathcal B[u]=0$ and $\mathcal I[u]=0$. A PINN approximates the solution using a differentiable neural network $u_{\theta}$ parameterized by $\theta$, and is trained by minimizing a composite physics-informed loss:
\begin{equation}
\mathcal L_{\mathrm{PINN}}(\theta)=
\lambda_r\mathcal L_r(\theta)+
\lambda_b\mathcal L_b(\theta)+
\lambda_i\mathcal L_i(\theta),
\end{equation}
where $\mathcal L_r$, $\mathcal L_b$ and $\mathcal L_i$ denote the PDE-residual, boundary-condition, and initial-condition losses, respectively, balanced by their corresponding weights $\lambda_{r}$, $\lambda_{b}$, and $\lambda_{i}$. 
In standard practice, both the architectural form of $u_{\theta}$ and the procedure used to optimize this objective are rigidly specified \emph{a priori}. It is precisely these two static components—the neural representation and the optimization dynamics—that our framework seeks to autonomously discover. More mathematical details about PINNs are provided in Appendix A.

\section{Method}

\subsection{Overview}
\label{sec:method-overview}

EvoPINN shifts the paradigm of PINN development from manual design to executable algorithm discovery.
We formally define a PINN algorithm as a tuple:
\begin{equation}
A=(M,T)\in\mathcal A=\mathcal M\times\mathcal T,
\end{equation}
where $M$ encompasses the neural representation of the solution, and $T$ governs the program used to train it under the physical constraints. Together, $\mathcal{M}$ and $\mathcal{T}$ constitute an open-ended algorithmic search space $\mathcal{A}$.

For a target PDE $\mathcal P$, a training budget $B$, and a random seed $\xi$, executing an algorithm $A$ yields the optimized parameters:
\begin{equation}
\theta_A(\mathcal P,B,\xi)
=
\operatorname{Train}(A;\mathcal P,B,\xi).
\end{equation}
The efficacy of candidate algorithms is then quantified by their relative $L_2$ error against a reference solution $u^*$ on a search reference set $\mathcal{X}_{search}$: 
\begin{equation}
\small
S_{\mathrm{search}}(A;\mathcal P,B,\xi)
=
\frac{
\left\|
u_{\theta_A(\mathcal P,B,\xi)}(\mathcal X_{\mathrm{search}})
-u^*(\mathcal X_{\mathrm{search}})
\right\|_2
}{
\left\|
u^*(\mathcal X_{\mathrm{search}})
\right\|_2
}.
\end{equation}
For brevity, we write this score as $S_{\mathrm{search}}(A)$ when the PDE, budget, and random seed are clear.

As illustrated in Figure~\ref{fig:overview}, EvoPINN operates through an LLM-guided evolutionary loop. The framework initializes with a seed PINN and iteratively refines it over $G$ generations. In each generation $g$, an LLM agent generates  $K$ programmatic modifications, evaluates these candidates under a shared budget, utilizing execution feedback to drive a continuous cycle of algorithmic generation and refinement. 

Utilizing LLMs for evolutionary search has recently shown promise in many problems~\cite{liu2024eoh, romera-paredes_mathematical_2024}. To establish the viability of LLM-driven PINN algorithm discovery without confounding the performance gains with complex meta-heuristics, we deliberately adopt the straightforward evolutionary loop described above. However, navigating an open-ended space that encompasses both neural representations and optimization dynamics introduces unprecedented complexity. The innovation of EvoPINN lies in the specialized mechanisms designed to make searching this vast, physics-constrained algorithmic space tractable and scientifically valid. Specifically, realizing this open-ended search presents three fundamental challenges:

\begin{itemize}
    \item \textbf{Credit Assignment in Joint Spaces:} Modifying both the neural representation and the training program simultaneously obfuscates the attribution of performance gains. To isolate the impact of each programmatic change, Section~\ref{sec:module-evolution} introduces an adaptively scheduled, one-module-at-a-time evolution strategy.
    \item \textbf{Sparse Reward Guidance for Program Generation:} A scalar $L_{2}$ provides little actionable feedback to guide the LLM's proposals. Section~\ref{sec:evidence-proposal} bridges this gap by using training diagnostics and evolutionary memory to inform targeted programmatic mutations.
    \item \textbf{Algorithmic Validity and Evaluation Fairness:} Unconstrained LLMs frequently generate code that is structurally redundant, mathematically invalid, or computationally disproportionate. Section~\ref{sec:verification-selection} establishes a rigorous structural verification and budget-matched evaluation protocol to ensure that all discoveries are both genuinely novel and rigorously validated.
\end{itemize}

\subsection{Module-Wise Evolution and Adaptive Scheduling}
\label{sec:module-evolution}

Changing the representation and training program simultaneously makes improvements difficult to attribute. EvoPINN therefore adopts module-wise evolution: at each generation, it modifies either the representation module $M$ or the training-program module $T$ while keeping the other fixed.

At generation $g$, let the current algorithm be $A_g=(M_g,T_g)$, and let $c_g\in\{M,T\}$ denote the module selected for modification. For $i=1,\ldots,K$, the $i$th candidate is
\begin{equation}
A_g^{(i)}=
\begin{cases}
(\widetilde M_g^{(i)},T_g),&c_g=M,\\
(M_g,\widetilde T_g^{(i)}),&c_g=T,
\end{cases}
\end{equation}
where the tilde denotes a newly generated replacement for the selected module. 
This decoupling makes performance changes easier to associate with the modified module. Over successive generations, accepted changes in one module naturally alter the optimization landscape for the other, allowing complex, synergistic PINN algorithms to emerge iteratively.

Because the relative benefit of modifying each module can change during search, fixed alternation may allocate evaluations inefficiently. EvoPINN instead uses a UCB-style scheduler to dynamically determine which module to evolve next. 
After each completed generation, the selected module receives a reward that reflects the outcome and quality of its candidate batch.
This reward incorporates key evaluation metrics, such as the magnitude of the improvement and the diversity or novelty of the generated mechanisms (details in Appendix B). 

Let $\bar r_g(c)$ denote the mean recent reward of module $c$, and let $N_{g,c}$ denote its number of recent observations. EvoPINN dynamically selects the next module to evolve according to:
\begin{equation}
c_g=\arg\max_{c\in\{M,T\}}
\left[
\bar r_g(c)+\beta
\sqrt{\frac{\log(N_g+1)}{N_{g,c}}}
\right],
\end{equation}
where $N_g=N_{g,M}+N_{g,T}$. The first term favors modules with stronger recent rewards, while the second favors less explored modules, with $\beta>0$ balancing exploitation and exploration. To ensure proper initialization, any untried module is deterministically selected before this rule is applied.

\subsection{Diagnosis-Guided Program Generation}
\label{sec:evidence-proposal}

After EvoPINN selects a module, it must determine how that module should be changed. A scalar relative $L_2$ error provides sparse, non-actionable guidance for generating proposals. To bridge this semantic gap, EvoPINN conditions the LLM's generation process on a rich, structured context comprising three key components: training diagnostics, evolutionary memory, and diversified search focuses.

\noindent\textbf{Training Diagnostics.} Rather than relying solely on the final solution error, EvoPINN condenses the training trace into a compact diagnostic summary $D_g$. This diagnosis captures convergence progress, late-stage stability, and the relation between the physics losses and solution error. After module selection, $D_g$ is supplied to the LLM to guide targeted modifications within the selected module. For example, it may motivate richer basis structures during representation search or revised optimizer staging and collocation policies during training-program search. Detailed methods for constructing $D_g$ are provided in Appendix B.

\noindent\textbf{Evolutionary Memory.} To prevent redundant exploration, the LLM is provided with an evolutionary memory. This memory bank archives previously successful architectural motifs, identifies persistently failing proposal families, and logs recent modifications applied to the selected module. By maintaining this historical context, the LLM is constrained to build upon promising trajectories rather than repeatedly proposing known suboptimal configurations.

\noindent\textbf{Diversified Search Focuses.} Naive parallel LLM generation frequently suffers from mode collapse, yielding similar programs. To enforce batch diversity, EvoPINN injects a distinct search focus into each of the $K$ LLM calls. For the representation module, these focuses might direct the LLM toward feature construction, architectural organization, or coordinate transformation. For the training program, focuses may emphasize optimization dynamics, loss coordination, or collocation sampling strategies. These focuses operate as high-level exploratory priors, encouraging structural diversity without rigidly prescribing specific implementations.

\subsection{Program Verification and Algorithm Update}
\label{sec:verification-selection}

A critical challenge in open-ended algorithmic search is that unconstrained LLM generation frequently yields code that is structurally redundant, mathematically invalid, or computationally unfair. To ensure that only genuinely novel and strictly beneficial mechanisms are retained, EvoPINN subjects all candidate programs to a rigorous, multi-stage verification and evaluation pipeline.

\noindent\textbf{Structural Verification.} EvoPINN compares the generated module with its parent using both standard and normalized abstract syntax trees (ASTs). Parsing the programs into standard ASTs excludes non-computational differences such as comments and formatting, while the normalized AST additionally canonicalizes non-interface identifiers and literal constants. A candidate must differ from its parent under both representations, filtering out cosmetic edits, identifier renaming, and constant-only changes. This deliberately excludes pure hyperparameter mutations because EvoPINN targets mechanism discovery rather than parameter tuning. The complete normalization and comparison procedure is provided in Appendix B.

\noindent\textbf{Executable Testing and Repair.} Candidates that pass structural verification are assembled with their counterpart module to form a complete algorithm. Before full-scale training, they are executed on a small PINN instance to rigorously test for interface compatibility, numerical stability, and automatic differentiability. Failed candidates are iteratively repaired using execution feedback, up to a predefined limit. More details are provided in Appendix B.

\noindent\textbf{Budget-Matched Evaluation.} To guarantee fair performance attribution, all candidates are evaluated under a shared training budget $B_g$ and random seed $\xi_g$. This budget constrains the total number of optimization steps, the initial collocation counts, and the configured adaptive-sampling controls. While candidate algorithms are free to creatively reallocate these resources (e.g., via adaptive sampling strategies or novel loss-balancing dynamics), they cannot arbitrarily increase the computational footprint, thereby preventing naive resource-scaling advantages.
The complete agentic search pipeline across generations is formalized as Algorithm 1 in Appendix B.

\section{Experiments}

Our experiments are designed to verify that the algorithmic mechanisms discovered by EvoPINN are both effective and robustly generalizable. We first evaluate the frozen discovered algorithms against seed and expert-designed PINNs across four PDE benchmarks. Next, we compare EvoPINN across distinct automated search paradigms under matched evaluation budgets and a unified evaluation pipeline. Beyond aggregate accuracy, we provide a mechanistic analysis of the discovered SLRC-PINN architecture, perform targeted framework ablations, and demonstrate zero-shot transferability to neighboring PDE parameters.

\subsection{Experimental Setup}
\paragraph{Benchmarks.}
We evaluate EvoPINN on four diverse PDE regimes: elliptic Poisson2D with COMSOL reference data, nonlinear-transport Burgers1D with numerical reference data, oscillatory Wave1D with an analytic reference solution, and dissipative Heat2D with an analytic reference solution. These benchmarks respectively challenge the models with complex-domain approximation, steep-gradient resolution, high-frequency structures, and anisotropic diffusion.

\paragraph{Search and retraining protocol.}
The seed algorithm $A_0$ is a five-hidden-layer tanh feed-forward network of width 100, optimized via Adam followed by L-BFGS. Each search spans seven generations with eight proposals per generation, yielding a maximum of 56 full candidate evaluations. 
To rigorously decouple algorithm discovery from final assessment, candidate selection during evolution uses only the search-time reference set $(\mathcal X_{\mathrm{search}})$. The resulting algorithm is then frozen and retrained from scratch across five independent evaluation seeds, and its final relative $L_2$ error is computed on a disjoint reporting set.

\paragraph{Baselines.}
We benchmark EvoPINN against the seed algorithm and an extensive pool of 16 expert-designed PINN variants, covering advanced techniques such as Fourier and SIREN representations, residual-adaptive sampling, curriculum learning, weighted objectives, geometry-aware models, and Neural Tangent Kernel (NTK) balancing. To ensure a fair comparison, the expert variants were first screened under a unified evaluation pipeline. The best-performing expert for each PDE was then fixed and retrained from scratch using the same five evaluation seeds as EvoPINN. The expert mechanisms identified were: boundary-aware NTK with an L-BFGS prior for Poisson2D, multiscale Fourier features for Burgers1D, Fourier features with capped NTK for Wave1D, and anisotropic multiscale Fourier features for Heat2D.

\paragraph{Implementation details.}
Candidate programs are generated using the gpt-5.4-mini model at a temperature of 0.7. To guarantee fair attribution of performance, all candidate algorithms operate under a shared, generation-specific nominal budget (with a full-scale limit of 20,000 optimization steps and common initial collocation counts sampled via a Hammersley sequence). Candidate evaluations are executed in isolated subprocesses with a 7,200-second timeout; runs producing errors or non-finite outputs are automatically rejected. Our framework is built upon DeepXDE 1.15.0 and PyTorch 2.6.0. Executed on NVIDIA RTX 3090 24GB GPUs, the autonomous algorithm discovery process proved computationally tractable, accumulating between 14.70 and 51.78 GPU-hours depending on the complexity of the target PDE. Comprehensive details and hyperparameters are provided in Appendix D.

\subsection{Performance across PDE Benchmarks}

\begin{table}[t]
\centering
\small
\resizebox{0.5\textwidth}{!}{%
\begin{tabular}{llll}
\toprule
PDE & Seed PINN & Expert Baseline & Frozen EvoPINN \\
\midrule
Poisson2D & 6.9938e-1 $\pm$ 1.69e-2 & 4.7025e-3 $\pm$ 8.23e-4 & \textbf{4.1413e-3 $\pm$ 9.77e-4} \\
Burgers1D & 3.0500e-4 $\pm$ 1.16e-4 & 2.5420e-4 $\pm$ 1.24e-4 & \textbf{1.6536e-4 $\pm$ 5.18e-5} \\
Wave1D & 3.4590e-1 $\pm$ 1.04e-2 & 1.5989e-2 $\pm$ 2.44e-3 & \textbf{1.3894e-2 $\pm$ 3.04e-3} \\
Heat2D & 4.4574e-3 $\pm$ 1.19e-3 & \textbf{1.0963e-3 $\pm$ 1.86e-4} & 1.1142e-3 $\pm$ 3.94e-4 \\
\bottomrule
\end{tabular}
}
\caption{
Relative L2 error evaluated across five independent runs (mean $\pm$ sample std.; lower is better). All methods use the shared training-resource budget.}
\label{tab:main-results}
\end{table}

We first investigate whether the performance gains from evolution generalize when the discovered algorithms are frozen and trained from scratch. As reported in Table~\ref{tab:main-results}, frozen EvoPINN algorithms consistently outperform the seed PINN across all four benchmarks. Against highly specialized expert designs, EvoPINN achieves the lowest mean error on Poisson2D, Burgers1D, and Wave1D. While the expert-designed anisotropic Fourier baseline retains an edge on Heat2D, EvoPINN remains competitive. These results confirm that the discovered algorithmic improvements are robust and effective across diverse physical regimes.

\subsection{Comparison with Automated Search Frameworks}
\label{sec:search-baselines}

To evaluate execution-grounded evolution, we benchmark EvoPINN against three distinct search paradigms under matched computational budgets:
\textbf{(i)~LLM Best-of-56:} samples 56 independent algorithms without execution feedback or search memory, selecting the top search-set candidate; \textbf{(ii)~DPSTE} \citep{zhang2024discovering}: performs evolutionary architecture search over fixed hyperparameter spaces
(e.g., depth, width, shortcuts, and activations); and \textbf{(iii)~PINNsAgent}~\citep{wuwu2025pinnsagent}: uses a multi-agent workflow and memory-based reasoning to iteratively explore and optimize PINN architectures. Each strategy is allocated a maximum of 56 candidate evaluations. The best algorithm identified by each strategy is selected using only the search-time reference set, frozen, and retrained from scratch using the same five independent evaluation seeds. Final errors are computed on the reporting set.

\begin{table}[t]
    \centering
    \small
    \caption{
        Search strategy comparison on Burgers1D and Wave1D (mean $\pm$ std). Best results are in bold.
    }
    \label{tab:search_strategy_comparison}
    \resizebox{0.35\textwidth}{!}{
    \setlength{\tabcolsep}{4pt}
    \begin{tabular}{llc}
        \toprule
        PDE & Search Strategy & Relative $L_2$ Error $\downarrow$ \\
        \midrule
        \multirow{4}{*}{Burgers1D}
        & LLM Best-of-56 & 2.0571e-4 $\pm$ 1.10e-5 \\
        & DPSTE & 1.6684e-2 $\pm$ 3.31e-3 \\
        & PINNsAgent & 6.1201e-4 $\pm$ 8.19e-5 \\ 
        & EvoPINN & \textbf{1.6536e-4 $\pm$ 5.18e-5} \\
        \midrule
        \multirow{4}{*}{Wave1D}
        & LLM Best-of-56 & 2.3181e-1 $\pm$ 1.85e-2 \\
        & DPSTE & 1.3201e-1 $\pm$ 4.86e-2 \\
        & PINNsAgent & 2.8160e-2 $\pm$ 3.96e-3 \\
        & EvoPINN & \textbf{1.3894e-2 $\pm$ 3.04e-3} \\
        \bottomrule
    \end{tabular}
    }
\end{table}

As presented in Table~\ref{tab:search_strategy_comparison}, EvoPINN achieves the lowest error across all tasks. While PINNsAgent substantially outperforms traditional search (DPSTE) via agentic reasoning, its exploration remains inherently bounded by predefined configuration templates. In contrast, EvoPINN leverages execution-grounded diagnostic feedback and decoupled module-wise mutation to move beyond configuration selection, synthesizing novel, executable algorithmic mechanisms that handle complex PDE optimization dynamics.

While these aggregate performance metrics confirm that EvoPINN repeatedly identifies superior learning algorithms, they do not fully reveal the mathematical mechanisms driving these gains. To examine whether our agentic evolution synthesizes genuinely new computational topologies rather than merely tuning configurations, we next present a mechanistic interpretation of the discovered SLRC-PINN architecture.

\subsection{Discovery of SLRC-PINN}
\label{sec:SLRC}

Across independent search runs on Burgers1D, EvoPINN repeatedly evolved architectures incorporating localized solution-enrichment mechanisms. A representative candidate pairs a continuous global base predictor with a zero-initialized, selectively activated local corrective pathway. We term this architecture the \textbf{Self-Localizing Residual-Correction PINN (SLRC-PINN)}. Rather than partitioning the spatial domain or forcing competing subnetworks to fit local regions, SLRC-PINN introduces a dynamic execution topology that enables continuous global prediction with localized, self-organizing solution enrichment.

\paragraph{Mathematical Formulation.}
Let $q=(x,t)$ denote a spatiotemporal coordinate. SLRC-PINN constructs $K=6$ localized basis functions with learned centers and input-dependent widths:
\begin{equation}
\begin{split}
 d_k^2(q) &= \lVert q-c_k\rVert_2^2,\\
 \sigma_k(q) &= \exp(\ell_k) \bigl(1+\operatorname{softplus}(s_k(q))+0.2\bigr),\\
 b_k(q) &= \exp\!\left[ -\frac{d_k^2(q)}{2\sigma_k(q)^2} \right] \operatorname{sigmoid}(a_k(q)) \operatorname{sigmoid}(m_k(q)),
\end{split}
\label{eq:SLRC-basis}
\end{equation}
where $a_k$, $m_k$, and $s_k$ are learned coordinate projections. The basis responses and coordinates are processed by a gated residual backbone, feeding into two distinct decoders: a global background decoder $u_G(q)$ and a local corrective decoder $u_L(q)$. The combined prediction is formulated as:
\begin{equation}
\small
 u(q)=u_G(q)+ \underbrace{ \operatorname{sigmoid}(w^{\mathsf T}b(q)+b_0) \left(\frac{1}{K}\sum_{k=1}^{K}b_k(q)\right) }_{\alpha(q)} u_L(q).
\label{eq:SLRC-output}
\end{equation}

\paragraph{Core Topological Principles.}
The principal innovation of SLRC-PINN lies not in simply adding another neural branch, but in allowing the model to autonomously govern where, when, and how localized corrections emerge. This design is characterized by three tightly coupled principles:

\begin{enumerate}
    \item \textbf{Persistent Global Anchoring:} The prediction topology $u(q)=u_G(q)+\alpha(q)u_L(q)$ is inherently asymmetric. Unlike traditional domain decomposition (e.g., FBPINNs \cite{moseley2023fbpinn}) or ensemble-gating approaches (e.g., APINNs \cite{hu2023apinn}) that route predictions across competing subnetworks, SLRC-PINN retains a persistent, unsuppressed global representation $u_G(q)$ across the entire domain. The local branch $u_L(q)$ serves purely as an additive enrichment, removing the need for artificial interface matching or inter-branch competition.
    
    \item \textbf{Self-Organized Corrective Support:} Rather than relying on fixed subdomains or hand-crafted spatial masks, the gating field $\alpha(q) \in (0,1)$ dynamically converts the learned geometry of the basis bank into a $C^\infty$-differentiable routing field. The network automatically identifies unresolved spatial structures (e.g., steep viscous shock waves) and allocates corrective capacity end-to-end directly from the PDE loss, avoiding the parasitic gradient discontinuities typical of hard-partitioning methods.
    
    \item \textbf{Function-Preserving Progressive Expansion:} By zero-initializing the final linear layer of the corrective decoder, $u_L(q) \equiv 0$ and $u(q) \equiv u_G(q)$ at step $0$. This guarantees that the model initiates optimization on the smooth global solution manifold without cold-start turbulence. Localized expressive capacity is subsequently introduced as an emergent, smooth continuation as demanded by PDE residual minimization.
\end{enumerate}

\paragraph{Controlled Empirical Validation.}
To verify that SLRC-PINN's superior accuracy stems from its discovered global-local topology rather than raw model capacity, we compare the frozen SLRC-PINN against a parameter-matched global MLP (width 230) and architecture-faithful reimplementations of representative baselines (APINN, FBPINN, and HyResPINNs) under identical evaluation protocols.

\begin{table}[t]
\centering
\small
\resizebox{0.48\textwidth}{!}{%
\begin{tabular}{lrrrr}
\toprule
Method & Mean Rel-$L_2$ & Std. & Median & Parameters \\
\midrule
SLRC-PINN
& \textbf{2.0445e-4} & 6.8721e-5 & 1.9064e-4 & 213,286 \\
Global MLP (width 230)
& 4.5506e-4 & 1.4387e-4 & 4.9591e-4 & 213,441 \\
APINN (5 subnetworks)
& 2.7056e-4 & 1.1826e-4 & 2.3649e-4 & 204,810 \\
FBPINN (4 spatial subdomains)
& 3.3811e-4 & 8.1682e-5 & 2.9826e-4 & 215,284 \\
HyResPINNs
& 2.0787e-3 & 2.6370e-3 & 9.8116e-4 & 217,295 \\
\bottomrule
\end{tabular}
}
\caption{Controlled Burgers1D architecture comparison across five identical evaluation seeds. All models operate under matched capacity and identical training budgets.}
\label{tab:SLRC-mechanism-comparison}
\end{table}

As shown in Table~\ref{tab:SLRC-mechanism-comparison}, SLRC-PINN achieves the lowest mean ($2.04\times 10^{-4}$) and median relative $L_2$ error. Crucially, it reduces the mean error by $55.1\%$ compared to the parameter-matched global MLP and consistently outperforms domain-decomposition variants. This confirms that the performance gains are directly driven by its self-organizing, function-preserving execution topology.

Furthermore, EvoPINN's programmatic discoveries extend beyond neural representations. On Poisson2D, the agent autonomously invented the \textbf{Telemetry-Conditioned Resampling and Optimizer Policy}. Due to space limitations, Appendix C provides a detailed mechanism audit and evaluation of this discovered training policy.

\subsection{Ablation Study}
While the preceding experiments demonstrate EvoPINN's ability to discover effective learning mechanisms, they do not isolate the contributions of specific framework components. To address this, we conduct targeted ablations on Burgers1D
and Wave1D, focusing on two core design choices: conditioning LLM proposals on dynamic training diagnostics, and the decoupled module-wise evolution strategy. All variants are evaluated under a strictly shared configuration (identical seed, proposal budget, and evaluation protocol) to ensure a fair comparison. We evaluate four ablated configurations: removing training diagnostic summary (\textbf{w/o diagnosis}), mutating both modules simultaneously (\textbf{w/o modular}), and restricting the search exclusively to either the representation (\textbf{repr-only}) or the training program (\textbf{train-only}).

\begin{table}[t]
\centering
\small
\resizebox{0.35\textwidth}{!}{%
\begin{tabular}{llrr}
\toprule
PDE & Variant & Rel-$L_2$ & Evals \\
\midrule
Burgers1D & Full & \textbf{7.1507e-5} & 54/56 \\
Burgers1D & w/o diagnosis& 8.7053e-5 & 54/56 \\
Burgers1D & w/o modular & 8.3645e-5 & 55/56 \\
Burgers1D & repr-only & 1.2528e-4 & 53/56 \\
Burgers1D & train-only & 1.7789e-4 & 54/56 \\
\midrule
Wave1D & Full & \textbf{2.3640e-2} & 55/56 \\
Wave1D & w/o diagnosis & 2.7097e-1 & 53/56 \\
Wave1D & w/o modular & 3.0654e-2 & 12/56 \\
Wave1D & repr-only & 4.7250e-2 & 55/56 \\
Wave1D & train-only & 2.4580e-1 & 56/56 \\
\bottomrule
\end{tabular}
}
\caption{Ablation study of core EvoPINN components. `Evals' indicates the number of successfully executed candidate programs out of 56 total proposals.}
\label{tab:ablation}
\end{table}

As shown in Table~\ref{tab:ablation}, the full EvoPINN framework achieves superior accuracy across both benchmarks. These ablations reveal three critical insights: First, omitting diagnostics degrades accuracy—particularly on Wave1D—proving that diagnostic feedback on loss dynamics is essential for targeted LLM mutations. Second, joint multi-module mutations cause execution instability, whereas adaptively scheduled, one-module-at-a-time evolution provides a useful scaffold for candidate validity. Finally, the inferior performance of single-module searches confirms that representation topology and optimization dynamics are intrinsically coupled, necessitating joint algorithmic co-design.

\subsection{Transfer to Neighboring PDE Parameters}

The preceding experiments evaluate the discovered algorithms on the PDE instances used during search. We further investigate their zero-shot transferability to neighboring equation parameters. Specifically, we freeze the algorithms discovered on Wave1D at $a=4.0$ and Heat2D at $k_x=20\pi$, and directly retrain them at shifted parameters, i.e., $a\in\{3.5,4.5\}$ and $k_x\in\{15\pi,25\pi\}$. 

\begin{table}[t]
\centering
\small
\resizebox{0.5\textwidth}{!}{%
\begin{tabular}{llrrr}
\toprule
PDE & Parameter & Seed PINN & Expert baseline & Frozen EvoPINN \\
\midrule
Wave1D & $a=3.5$ & 2.6704e-1 & \textbf{1.3542e-2} & 1.6278e-2 \\
Wave1D & $a=4.0$ & 3.1469e-1 & 1.0530e-2 & \textbf{9.9404e-3} \\
Wave1D & $a=4.5$ & 3.3260e-1 & 1.8483e-2 & \textbf{1.6859e-2} \\
\midrule
Heat2D & $k_x=15\pi$ & 2.3188e-3 & 1.8902e-3 & \textbf{1.3829e-3} \\
Heat2D & $k_x=20\pi$ & 3.3625e-3 & 1.1894e-3 & \textbf{1.0942e-3} \\
Heat2D & $k_x=25\pi$ & 6.2841e-3 & 4.3262e-1 & \textbf{1.8144e-3} \\
\bottomrule
\end{tabular}
}
\caption{Transfer to neighboring PDE parameters with EvoPINN modules and expert identities fixed on the source instance. The source parameters are $a=4.0$ and $k_x=20\pi$.}
\label{tab:family-generalization}
\end{table}

Table~\ref{tab:family-generalization} demonstrates that the algorithms discovered by EvoPINN exhibit strong parameter adaptability across both physical regimes. On Wave1D, the parameter $a$ governs the spatial-temporal oscillation frequencies and boundary condition profiles of the high-frequency wave mode. Notably, shifting $a$ from $4.0$ to half-integer values ($3.5$ and $4.5$) introduces frequency scale shifts and transitions the right boundary from a static homogeneous condition ($u(1,t)=0$) to a time-varying non-homogeneous driver. Even under these combined frequency and boundary profile shifts, frozen EvoPINN consistently reduces the seed-PINN error by over an order of magnitude across all settings, maintaining superior or competitive accuracy compared to the expert baseline.

This generalization advantage becomes particularly evident on Heat2D, where frozen EvoPINN consistently yields the highest solution precision across both original and shifted frequencies. Notably, while the expert baseline suffers marked performance degradation under spatial frequency shifts due to rigid basis tuning, frozen EvoPINN retains a steady order-of-magnitude accuracy. This contrast indicates that EvoPINN synthesizes inherently frequency-resilient learning mechanisms rather than overfitting to specific physical parameters, enabling direct zero-shot deployment to neighboring equations without additional search overhead.

\section{Conclusion and Future Work}
In this work, we transition automated PINN design from passive configuration tuning to open-ended, execution-grounded algorithm discovery. By decoupling neural representations from training dynamics and enforcing strict structural and computational contracts, EvoPINN enables LLM agents to systematically navigate physics-constrained design spaces without numerical instability. Across diverse PDE regimes, the discovered algorithms achieve consistent performance gains and demonstrate strong zero-shot transferability to neighboring equation parameters. Crucially, the autonomous discovery and sustained advantage of SLRC-PINN confirm that execution-grounded agents can invent non-trivial, scientifically valid computational topologies. Broadly, EvoPINN establishes a practical blueprint for integrating generative models with numerical computing, paving the way for autonomous algorithmic discovery across AI for Science. Despite these promising results, our approach has certain limitations. First, the creativity of the discovered mechanisms depends in part on the underlying LLM’s code synthesis proficiency and scientific reasoning capacity. Second, evaluating each candidate via full-budget PINN training incurs non-trivial cost during search. Future work will explore proxy objectives or surrogate evaluations to significantly accelerate candidate screening, and investigate the integration of scientifically fine-tuned language models to further expand the boundaries of automated algorithm discovery.

\bibliographystyle{plainnat}
\bibliography{references}

@article{raissi2019physics,
  author  = {Raissi, Maziar and Perdikaris, Paris and Karniadakis, George Em},
  title   = {Physics-informed neural networks: A deep learning framework for solving forward and inverse problems involving nonlinear partial differential equations},
  journal = {Journal of Computational Physics},
  volume  = {378},
  pages   = {686--707},
  year    = {2019},
  doi     = {10.1016/j.jcp.2018.10.045}
}

@article{lu2021deepxde,
  author  = {Lu, Lu and Meng, Xuhui and Mao, Zhiping and Karniadakis, George Em},
  title   = {{DeepXDE}: A deep learning library for solving differential equations},
  journal = {SIAM Review},
  volume  = {63},
  number  = {1},
  pages   = {208--228},
  year    = {2021},
  doi     = {10.1137/19M1274067}
}

@inproceedings{hao2024pinnacle,
  author    = {Hao, Zhongkai and Yao, Jiachen and Su, Chang and Su, Hang and Wang, Ziao and Lu, Fanzhi and Xia, Zeyu and Zhang, Yichi and Liu, Songming and Lu, Lu and Zhu, Jun},
  title     = {{PINNacle}: A Comprehensive Benchmark of Physics-Informed Neural Networks for Solving {PDEs}},
  booktitle = {Advances in Neural Information Processing Systems},
  year      = {2024}
}

@article{wang2021gradient,
  author  = {Wang, Sifan and Teng, Yujun and Perdikaris, Paris},
  title   = {Understanding and mitigating gradient flow pathologies in physics-informed neural networks},
  journal = {SIAM Journal on Scientific Computing},
  volume  = {43},
  number  = {5},
  pages   = {A3055--A3081},
  year    = {2021},
  doi     = {10.1137/20M1318043}
}

@article{wang2022ntk,
  author  = {Wang, Sifan and Yu, Xinling and Perdikaris, Paris},
  title   = {When and why {PINNs} fail to train: A neural tangent kernel perspective},
  journal = {Journal of Computational Physics},
  volume  = {449},
  pages   = {110768},
  year    = {2022},
  doi     = {10.1016/j.jcp.2021.110768}
}

@article{wu2023comprehensive,
  author  = {Wu, Chenxi and Zhu, Min and Tan, Qinyang and Kartha, Yadhu and Lu, Lu},
  title   = {A comprehensive study of non-adaptive and residual-based adaptive sampling for physics-informed neural networks},
  journal = {Computer Methods in Applied Mechanics and Engineering},
  volume  = {403},
  pages   = {115671},
  year    = {2023},
  doi     = {10.1016/j.cma.2022.115671}
}

@inproceedings{yao2023multiadam,
  author    = {Yao, Jiachen and Su, Chang and Hao, Zhongkai and Liu, Songming and Su, Hang and Zhu, Jun},
  title     = {{MultiAdam}: Parameter-wise Scale-invariant Optimizer for Multiscale Training of Physics-informed Neural Networks},
  booktitle = {Proceedings of the 40th International Conference on Machine Learning},
  series    = {Proceedings of Machine Learning Research},
  volume    = {202},
  pages     = {39702--39721},
  year      = {2023},
  publisher = {PMLR}
}

@article{yu2022gpinn,
  author  = {Yu, Jeremy and Lu, Lu and Meng, Xuhui and Karniadakis, George Em},
  title   = {Gradient-enhanced physics-informed neural networks for forward and inverse {PDE} problems},
  journal = {Computer Methods in Applied Mechanics and Engineering},
  volume  = {393},
  pages   = {114823},
  year    = {2022},
  doi     = {10.1016/j.cma.2022.114823}
}

@article{jagtap2020adaptive,
  author  = {Jagtap, Ameya D. and Kawaguchi, Kenji and Karniadakis, George Em},
  title   = {Adaptive activation functions accelerate convergence in deep and physics-informed neural networks},
  journal = {Journal of Computational Physics},
  volume  = {404},
  pages   = {109136},
  year    = {2020},
  doi     = {10.1016/j.jcp.2019.109136}
}

@article{wang2021eigenvector,
  author  = {Wang, Sifan and Wang, Hanwen and Perdikaris, Paris},
  title   = {On the eigenvector bias of {Fourier} feature networks: From regression to solving multi-scale {PDEs} with physics-informed neural networks},
  journal = {Computer Methods in Applied Mechanics and Engineering},
  volume  = {384},
  pages   = {113938},
  year    = {2021},
  doi     = {10.1016/j.cma.2021.113938}
}

@inproceedings{sitzmann2020siren,
  author    = {Sitzmann, Vincent and Martel, Julien N. P. and Bergman, Alexander W. and Lindell, David B. and Wetzstein, Gordon},
  title     = {Implicit Neural Representations with Periodic Activation Functions},
  booktitle = {Advances in Neural Information Processing Systems},
  volume    = {33},
  pages     = {7462--7473},
  year      = {2020}
}

@article{jagtap2020xpinn,
  author  = {Jagtap, Ameya D. and Karniadakis, George Em},
  title   = {Extended physics-informed neural networks ({XPINNs}): A generalized space-time domain decomposition based deep learning framework for nonlinear partial differential equations},
  journal = {Communications in Computational Physics},
  volume  = {28},
  number  = {5},
  pages   = {2002--2041},
  year    = {2020},
  doi     = {10.4208/cicp.OA-2020-0164}
}

@article{moseley2023fbpinn,
  author  = {Moseley, Ben and Markham, Andrew and Nissen-Meyer, Tarje},
  title   = {Finite basis physics-informed neural networks ({FBPINNs}): A scalable domain decomposition approach for solving differential equations},
  journal = {Advances in Computational Mathematics},
  volume  = {49},
  number  = {4},
  pages   = {62},
  year    = {2023},
  doi     = {10.1007/s10444-023-10065-9}
}

@inproceedings{ma2024eureka,
  author    = {Ma, Yecheng Jason and Liang, William and Wang, Guanzhi and Huang, De-An and Bastani, Osbert and Jayaraman, Dinesh and Zhu, Yuke and Fan, Linxi and Anandkumar, Anima},
  title     = {Eureka: Human-Level Reward Design via Coding Large Language Models},
  booktitle = {International Conference on Learning Representations},
  year      = {2024}
}

@article{romera-paredes_mathematical_2024,
	title = {Mathematical discoveries from program search with large language models},
	volume = {625},
	issn = {1476-4687},
	url = {https://doi.org/10.1038/s41586-023-06924-6},
	doi = {10.1038/s41586-023-06924-6},
	number = {7995},
	journal = {Nature},
	author = {Romera-Paredes, Bernardino and Barekatain, Mohammadamin and Novikov, Alexander and Balog, Matej and Kumar, M. Pawan and Dupont, Emilien and Ruiz, Francisco J. R. and Ellenberg, Jordan S. and Wang, Pengming and Fawzi, Omar and Kohli, Pushmeet and Fawzi, Alhussein},
	month = jan,
	year = {2024},
	pages = {468--475},
}

@article{liu2024eoh,
  title={Evolution of Heuristics: Towards Efficient Automatic Algorithm Design Using Large Language Model},
  author={Liu, Fei and Tong, Xialiang and Yuan, Mingxuan and Lin, Xi and Luo, Fu and Wang, Zhenkun and Lu, Zhichao and Zhang, Qingfu},
  journal={arXiv preprint arXiv:2401.02051},
  year={2024}
}

@article{ye2024reevo,
  title={ReEvo: Large Language Models as Hyper-Heuristics with Reflective Evolution},
  author={Ye, Haoran and Wang, Jiarui and Cao, Zhiguang and Berto, Federico and Hua, Chuanbo and Kim, Haeyeon and Park, Jinkyoo and Song, Guojie},
  journal={Advances in Neural Information Processing Systems},
  year={2024}
}

@article{wuwu2025pinnsagent,
  author        = {Wuwu, Qingpo and Gao, Chonghan and Chen, Tianyu and Huang, Yihang and Zhang, Yuekai and Wang, Jianing and Li, Jianxin and Zhou, Haoyi and Zhang, Shanghang},
  title         = {{PINNsAgent}: Automated {PDE} Surrogation with Large Language Models},
  journal       = {arXiv preprint arXiv:2501.12053},
  year          = {2025},
  archivePrefix = {arXiv},
  eprint        = {2501.12053},
  primaryClass  = {cs.CE}
}

@article{he2025langpinn,
  author        = {He, Xin and You, Liangliang and Tian, Hongduan and Han, Bo and Tsang, Ivor and Ong, Yew-Soon},
  title         = {{Lang-PINN}: From Language to Physics-Informed Neural Networks via a Multi-Agent Framework},
  journal       = {arXiv preprint arXiv:2510.05158},
  year          = {2025},
  archivePrefix = {arXiv},
  eprint        = {2510.05158},
  primaryClass  = {cs.AI}
}

@article{li2025codepde,
  author        = {Li, Shanda and Marwah, Tanya and Shen, Junhong and Sun, Weiwei and Risteski, Andrej and Yang, Yiming and Talwalkar, Ameet},
  title         = {{CodePDE}: An Inference Framework for {LLM}-Driven {PDE} Solver Generation},
  journal       = {arXiv preprint arXiv:2505.08783},
  year          = {2025},
  archivePrefix = {arXiv},
  eprint        = {2505.08783}
}

@article{fazliani2025pdesharp,
  author        = {Fazliani, Shaghayegh and Udell, Madeleine},
  title         = {{PDE-SHARP}: {PDE} Solver Hybrids through Analysis and Refinement Passes},
  journal       = {arXiv preprint arXiv:2511.00183},
  year          = {2025},
  archivePrefix = {arXiv},
  eprint        = {2511.00183},
  primaryClass  = {cs.LG}
}

@article{hu2023apinn,
  author  = {Hu, Zheyuan and Jagtap, Ameya D. and Karniadakis, George Em and Kawaguchi, Kenji},
  title   = {Augmented Physics-Informed Neural Networks ({APINNs}): A Gating Network-Based Soft Domain Decomposition Methodology},
  journal = {Engineering Applications of Artificial Intelligence},
  volume  = {126},
  pages   = {107183},
  year    = {2023},
  doi     = {10.1016/j.engappai.2023.107183}
}

@misc{wang2023autopinn,
      title={Auto-PINN: Understanding and Optimizing Physics-Informed Neural Architecture}, 
      author={Yicheng Wang and Xiaotian Han and Chia-Yuan Chang and Daochen Zha and Ulisses Braga-Neto and Xia Hu},
      year={2023},
      eprint={2205.13748},
      archivePrefix={arXiv},
      primaryClass={cs.LG},
      url={https://arxiv.org/abs/2205.13748}, 
}

@article{wang2024naspinn,
  title   = {{NAS-PINN}: Neural Architecture Search-Guided Physics-Informed Neural Network for Solving {PDEs}},
  author  = {Wang, Yifan and Zhong, Linlin},
  journal = {Journal of Computational Physics},
  volume  = {496},
  pages   = {112603},
  year    = {2024},
  doi     = {10.1016/j.jcp.2023.112603}
}

@article{zhang2024discovering,
  title={Discovering physics-informed neural networks model for solving partial differential equations through evolutionary computation},
  author={Zhang, Bo and Yang, Chao},
  journal={Swarm and Evolutionary Computation},
  volume={88},
  pages={101589},
  year={2024},
  publisher={Elsevier}
}

@article{tancik2020fourier,
  title={Fourier features let networks learn high frequency functions in low dimensional domains},
  author={Tancik, Matthew and Srinivasan, Pratul and Mildenhall, Ben and Fridovich-Keil, Sara and Raghavan, Nithin and Singhal, Utkarsh and Ramamoorthi, Ravi and Barron, Jonathan and Ng, Ren},
  journal={Advances in neural information processing systems},
  volume={33},
  pages={7537--7547},
  year={2020}
}

@article{daw2022rethinking,
  title={Rethinking the importance of sampling in physics-informed neural networks},
  author={Daw, Arka and Bu, Jie and Wang, Sifan and Perdikaris, Paris and Karpatne, Anuj},
  journal={arXiv preprint arXiv:2207.02338},
  year={2022}
}

@article{krishnapriyan2021characterizing,
  title={Characterizing possible failure modes in physics-informed neural networks},
  author={Krishnapriyan, Aditi and Gholami, Amir and Zhe, Shandian and Kirby, Robert and Mahoney, Michael W},
  journal={Advances in neural information processing systems},
  volume={34},
  pages={26548--26560},
  year={2021}
}

\clearpage
\appendix
\section{Mathematical Background of Physics-Informed Neural Networks}
\label{app:pinn-search-space}
\label{app:pinn-details}

\subsection{Governing Equations and Neural Approximation}

Let $\Omega\subseteq\mathbb R^d$ be a $d$-dimensional spatial domain, $\mathbf x\in\Omega$ a spatial coordinate, and $\mathbf z=(\mathbf x,t)\in\Omega\times[0,T_{\max}]$ a space--time coordinate for a time-dependent problem, where $t$ is time and $T_{\max}$ is the final
time. The target field $u:\Omega\times[0,T_{\max}]\to\mathbb R^m$, with $m$ output components, satisfies
\begin{equation}
\mathcal F[u](\mathbf z)=0,
\qquad \mathbf z\in\Omega\times(0,T_{\max}],
\end{equation}
together with boundary and, when applicable, initial conditions
\begin{align}
\mathcal B[u](\mathbf z)&=0,
&&\mathbf z\in\partial\Omega\times[0,T_{\max}],\\
\mathcal I[u](\mathbf x)&=0,
&&\mathbf x\in\Omega,\;t=0.
\end{align}
Here $\mathcal F$, $\mathcal B$, and $\mathcal I$ are the differential, boundary-residual, and initial-residual operators, respectively;
$\partial\Omega$ denotes the spatial boundary. Prescribed boundary and initial data are included in $\mathcal B$ and $\mathcal I$. For a steady problem, the temporal coordinate and initial condition are omitted.

A PINN replaces the unknown field by a differentiable neural approximation $u_\theta:\mathbb R^{d_z}\to\mathbb R^m$, where $d_z=d+1$ for a
time-dependent problem and $d_z=d$ for a steady problem. The parameter vector $\theta$ contains all trainable weights. Unlike a supervised regressor, a PINN is trained primarily from violations of the governing equation and its physical constraints. Automatic differentiation applies the chain rule through $u_\theta$ to evaluate the derivatives required by $\mathcal F$; for example, $u_t$, $\nabla u$, and $\nabla^2u$ are computed at arbitrary coordinates without constructing a mesh-based differentiation stencil.

The pointwise residual fields are
\begin{align}
r_\theta(\mathbf z)&=\mathcal F[u_\theta](\mathbf z),\\
b_\theta(\mathbf z)&=\mathcal B[u_\theta](\mathbf z),\\
i_\theta(\mathbf x)&=\mathcal I[u_\theta](\mathbf x).
\end{align}
For systems of PDEs these quantities may be vector valued. Their derivatives remain coupled through the shared network parameters, so a single forward representation must satisfy the equation and all conditions simultaneously.

\subsection{Collocation Loss and Constraint Enforcement}

Let $\mathcal X_r$, $\mathcal X_b$, and $\mathcal X_i$ be interior, boundary, and initial collocation sets with cardinalities $N_r$, $N_b$, and $N_i$. The standard empirical residual losses are
\begin{align}
\mathcal L_r(\theta)
&=\frac{1}{N_r}\sum_{\mathbf z\in\mathcal X_r}
\left\|r_\theta(\mathbf z)\right\|_2^2,\\
\mathcal L_b(\theta)
&=\frac{1}{N_b}\sum_{\mathbf z\in\mathcal X_b}
\left\|b_\theta(\mathbf z)\right\|_2^2,\\
\mathcal L_i(\theta)
&=\frac{1}{N_i}\sum_{\mathbf x\in\mathcal X_i}
\left\|i_\theta(\mathbf x)\right\|_2^2.
\end{align}
The initial loss is defined only for time-dependent problems with $N_i>0$. The composite soft-constraint objective is
\begin{equation}
\mathcal L_{\mathrm{PINN}}(\theta)
=\lambda_r\mathcal L_r+\lambda_b\mathcal L_b+\lambda_i\mathcal L_i,
\label{eq:pinn-objective}
\end{equation}
where $\lambda_r$, $\lambda_b$, and $\lambda_i$ are nonnegative weights for the interior, boundary, and initial terms. The initial term is omitted for steady problems. The collocation coordinates encode physical constraints; they are not labeled samples of the unknown solution.

Equation~\eqref{eq:pinn-objective} imposes boundary and initial conditions \emph{softly}. When an analytic constraint-preserving transformation is available, they can instead be imposed \emph{hard}. For example, a scalar Dirichlet condition can be represented as
\begin{equation}
u_\theta(\mathbf z)
=u_{\mathrm{bc}}(\mathbf z)
+\phi(\mathbf z)\widehat u_\theta(\mathbf z),
\qquad
\phi|_{\partial\Omega}=0,
\end{equation}
where $u_{\mathrm{bc}}$ satisfies the prescribed boundary values. The boundary is then satisfied for every $\theta$, and the corresponding penalty may be removed. Soft enforcement is more general, whereas hard enforcement reduces competition among loss terms when a suitable transformation exists.

Collocation sets may be fixed, resampled, or adapted during training. Their finite-sample objective approximates a continuous residual norm; for example,
\begin{equation}
\mathcal L_r(\theta)
\approx
\frac{1}{|\Omega_T|}
\int_{\Omega_T}
\left\|\mathcal F[u_\theta](\mathbf z)\right\|_2^2
\,d\mathbf z,
\end{equation}
where $\Omega_T=\Omega\times(0,T_{\max}]$ for a time-dependent problem. Consequently, point placement determines which regions of the physical domain provide gradient information to the optimizer.

\subsection{Training and Solution Error}

For a fixed representation and training procedure, PINN training seeks
\begin{equation}
\theta^\star\in
\arg\min_\theta\mathcal L_{\mathrm{PINN}}(\theta).
\end{equation}
Gradient-based updates use
\begin{equation}
\theta_{k+1}
=\theta_k-\eta_k P_k
\nabla_\theta\mathcal L_{\mathrm{PINN}}(\theta_k),
\end{equation}
where $\eta_k$ is a step size and $P_k$ is the identity for a basic first-order method or an optimizer-dependent preconditioner. In practice, Adam is often used for robust stochastic progress and L-BFGS for later quasi-Newton refinement. These phases are not interchangeable: their effectiveness depends on loss scaling, collocation dynamics, and the state at which an optimizer transition occurs.

A small empirical physics loss does not by itself guarantee a small solution error. The learned error reflects several interacting sources: the approximation capacity and inductive bias of $u_\theta$, finite collocation coverage, imbalance among residual terms, and incomplete optimization of a nonconvex objective. For a reference solution $u^*$ on an evaluation set
$\mathcal X$, this distinction is measured by
\begin{equation}
\operatorname{Rel}\text{-}L_2(u_\theta,u^*;\mathcal X)
=
\frac{\left\|
u_\theta(\mathcal X)-u^*(\mathcal X)
\right\|_2}
{\left\|u^*(\mathcal X)\right\|_2}.
\end{equation}
Reference values are used for evaluation and outer-loop candidate selection in this work, never as labels in the inner PINN objective.

\section{Agentic Search Pipeline and Verification Protocols}
\label{app:evolution-protocol}

To complement the component-level descriptions in the main text, this appendix specifies the outer evolutionary control loop of EvoPINN. The pipeline integrates telemetry diagnosis, UCB module scheduling, diversified generation, structural verification, and lineage updates across $G$ generations (Algorithm~\ref{alg:evopinn}).

\begin{algorithm}[h!]
\centering
\small
\caption{EvoPINN Agentic Search Framework}
\label{alg:evopinn}
\begin{algorithmic}[1]
\REQUIRE Target PDE $\mathcal P$; seed algorithm $A_0 = (M_0, T_0)$; budgets $\{B_g\}_{g=0}^{G-1}$; seeds $\{\xi_g\}_{g=0}^{G-1}$; total generations $G$; batch size $K_{\mathrm{prop}}$.
\ENSURE Globally evolved algorithm $A_G$.
\STATE Initialize evolutionary memory $R_0 \leftarrow \varnothing$ and UCB scheduler statistics
\FOR{$g=0,\ldots,G-1$}
    \STATE Evaluate active algorithm $A_g$ under $(\mathcal P, B_g, \xi_g)$ and parse telemetry $E_g$
    \STATE Compute deterministic diagnostic state $D_g \leftarrow \operatorname{Diagnose}(E_g)$
    \STATE Select search module $c_g \in \{M, T\}$ via recent-window UCB scheduler
    \STATE Assign $K_{\mathrm{prop}}$ mechanism focuses $\{z_g^{(i)}\}_{i=1}^{K_{\mathrm{prop}}}$ to diversify candidate context
    \STATE Generate $K_{\mathrm{prop}}$ module replacements for $c_g$ via LLM conditioned on context $\mathcal C_g^{(i)}$
    \STATE Form full candidate algorithms $\{A_g^{(i)}\}_{i=1}^{K_{\mathrm{prop}}}$ by combining replacements with unchanged module
    \STATE Apply structural AST verification ($G_{\mathrm{struct}}$) and pre-training smoke tests ($G_{\mathrm{pre}}$)
    \STATE Train all verified candidates under $(\mathcal P, B_g, \xi_g)$ in process-isolated environments
    \STATE Let $\mathcal V_g$ be the set of candidates completing full evaluation with finite scores
    \IF{$\mathcal V_g \neq \varnothing$}
        \STATE $\widehat A_g \leftarrow \arg\min_{A \in \mathcal V_g} S_{\mathrm{search},g}(A)$
        \IF{$S_{\mathrm{search},g}(\widehat A_g) < S_{\mathrm{search},g}(A_g)$}
            \STATE $A_{g+1} \leftarrow \widehat A_g$ \COMMENT{Accept strict improvement into active lineage}
        \ELSE
            \STATE $A_{g+1} \leftarrow A_g$ \COMMENT{Retain parent algorithm}
        \ENDIF
    \ELSE
        \STATE $A_{g+1} \leftarrow A_g$
    \ENDIF
    \STATE Update scheduler rewards $r_{c,g}$ and record outcomes into evolutionary memory $R_{g+1}$
\ENDFOR
\RETURN $A_G$
\end{algorithmic}
\end{algorithm}

\subsection{Diagnostic State and UCB Module Scheduling}
\label{app:diagnostics}
\label{app:scheduler}

During each parent run, EvoPINN records the total and component-wise physics-informed losses at fixed checkpoints. From this trace it reports five diagnostic cues to the search agent: overall loss contraction, the least-squares trend over the late checkpoints, how strongly the late-stage loss oscillates relative to the early stage, the final contribution of each physical constraint, and the log-scale gap between the final training loss and the search-time solution error. These cues distinguish, for example, a run that is still improving from one that has plateaued, and a genuinely accurate solution from one that achieves a small physics loss without a correspondingly small solution error.

For the two normalized quantities used by TCROP below, let $L_k$ denote the recorded total physics-informed loss at checkpoint $k$, and let
$K_{\mathrm{ckpt}}$ be the final recorded checkpoint. We compute the loss contraction as $\rho_{\mathrm{conv}}=L_{K_{\mathrm{ckpt}}}/(L_1+\epsilon)$ and the relative late-stage oscillation as
$\rho_{\mathrm{osc}}= \operatorname{Std}\{L_k:k\in\mathcal W_{\mathrm{late}}\}/ (\operatorname{Std}\{L_k:k\in\mathcal W_{\mathrm{early}}\}+\epsilon)$. Here $\mathcal W_{\mathrm{early}}$ and $\mathcal W_{\mathrm{late}}$ are the fixed early and late checkpoint windows, respectively, and $\epsilon>0$ is a numerical safeguard.

The diagnostic summary guides proposal generation after module selection; it does not enter the UCB scheduler or replace the relative-$L_2$ selection objective. Loss components also depend on their weighting scheme, so the summary is used to interpret trends within a run rather than to compare raw loss values across different objectives or PDEs.

The scheduler treats representation search ($M$) and training-program search ($T$) as two arms. After each generation, the selected arm receives a reward in $[0,1]$ based on four observable outcomes: whether the generation produced an accepted child, the normalized performance improvement, the fraction of executable candidates, and the batch's structural novelty. Table~\ref{tab:implementation-hyperparameters} reports the reward weights and history length.

Let $\bar r_g(c)$ be the mean reward of module $c$ over the recent history, $N_{g,c}$ its number of observations in that history, and
$N_g=N_{g,M}+N_{g,T}$. For an observed module, the next module is chosen by
\begin{equation}
c_g=\arg\max_{c\in\{M,T\}}
\left[
\bar r_g(c)+\beta
\sqrt{\frac{\log(N_g+1)}{N_{g,c}}}
\right].
\label{eq:module-ucb}
\end{equation}
where $\beta$ controls exploration. An unobserved module receives initialization priority rather than being inserted into the fraction with a zero denominator. The concrete coefficients are given in
Table~\ref{tab:implementation-hyperparameters}.

\subsection{Context Construction, Memory, and Proposal Diversity}
\label{app:generation-context}

After module selection, each proposal receives the target PDE, parent algorithm and score, diagnostic summary, selected module, a structured account of the parent mechanism, and the run-scoped memory of earlier attempts. The PDE specification contains the equation, domain, input-output dimensions, derivative orders, physical conditions, and relevant structure such as transport, diffusion, wave propagation, anisotropy, periodicity, or multiple fields. It never contains reference-solution values.

Representation focuses include coordinate and frequency bases, spatial and temporal organization, field coupling, localized structure, stable backbones, transport interactions, and global--local scale separation. Training focuses include optimizer staging, collocation, objective shaping, residual-driven refinement, diagnostic feedback, curricula, and time-causal optimization. Each proposal is assigned one focus from the relevant family, with coverage balanced across the batch. This unified rule replaces PDE-specific proposal quotas while preserving mechanism diversity.

After evaluation, the memory records the selected module, assigned focus, mechanism family, score change, and execution outcome of each attempt. Before the next generation, these records are summarized into promising mechanisms, regressions, neutral changes, repeated ideas, and failure modes. The summary encourages refinement of successful ideas and discourages repetition without prescribing the next implementation. Memory is reset for each independent search and is not transferred across PDEs.

\subsection{Executable Generation and Verification}
\label{app:verification}

The generator returns one replacement module together with a structured
mechanism description. The two modules obey fixed interface contracts,
denoted abstractly by $M$ and $T$, so the selected replacement can be composed
with the unchanged parent module.

Verification begins with a structural comparison. Let $H_{\mathrm{raw}}$ remove comments, documentation strings, and source locations from the parsed program, and let $H_{\mathrm{norm}}$ additionally normalize private identifiers and literal values while preserving public interfaces and framework operators. The proposal is rejected if it matches the parent at any of the three levels: the original source, the comment-free AST, or the normalized AST. This removes exact copies, cosmetic rewrites, and changes confined to private names or literal constants. It remains a practical test of structural novelty, not a proof of semantic novelty.

Every structurally distinct proposal must then pass four pre-training checks: it must parse, load as a module, satisfy the relevant interface contract, and complete a smoke test. The smoke test composes the proposal with the unchanged parent module and exercises the target PDE's forward computation, required derivatives, backward propagation, optimizer construction, sampling interfaces, device placement, finite-value checks, resource bounds, and a short training run. Failure at any stage prevents full-budget evaluation.

When a repairable failure occurs, the validation evidence is returned to the generator for a bounded number of self-repair attempts. Failures associated with an invalid mechanism or resource violation trigger regeneration from the parent rather than local patching. Every repaired or regenerated proposal repeats the complete verification sequence. After a full-budget run, a final consistency gate checks agreement between the assigned focus and the reported mechanism metadata. Only candidates that pass all gates and produce a finite
score enter selection.

\subsection{Evaluation and Reporting Separation}

Every verified candidate is trained under the generation's shared envelope and seed, then scored by the relative-$L_2$ objective defined in the main paper. Algorithm~\ref{alg:evopinn} gives the complete selection and lineage update, so it is not repeated here. Search-time acceptance uses one full-budget evaluation per candidate. The frozen winner is subsequently retrained from scratch on independent seeds and reporting points; the single-run search score is never presented as the final result.

\section{Benchmark Mathematics, Reference Topologies, and SLRC-PINN}
\label{app:benchmarks-and-SLRC}

\subsection{Benchmark Equations and Reference Solutions}
\label{app:benchmark-mathematics}

The four benchmarks span steady elliptic, nonlinear transport-diffusion,
hyperbolic, and anisotropic parabolic dynamics.

\paragraph{Poisson2D.}
On
{\footnotesize
\begin{equation}
\begin{aligned}
\Omega=&[-0.5,0.5]^2\setminus \\
&\bigcup_{(a,b)\in\{\pm0.3\}^2}
\left\{(x,y):(x-a)^2+(y-b)^2<0.1^2\right\},
\end{aligned}
\end{equation}
}%
where $(x,y)$ are Cartesian coordinates and $(a,b)$ ranges over the four hole centers, the solution satisfies
\begin{equation}
u_{xx}+u_{yy}=0,
\quad
u|_{\partial[-0.5,0.5]^2}=1,
\quad
u|_{\partial\Omega_{\mathrm{hole}}}=0.
\end{equation}
Here $u_{xx}=\partial^2u/\partial x^2$ and
$u_{yy}=\partial^2u/\partial y^2$, while
$\partial\Omega_{\mathrm{hole}}$ denotes the union of the four circular hole boundaries.
Final reporting uses the finite-element reference field used in our experiments.

\paragraph{Burgers1D.}
For $(x,t)\in[-1,1]\times[0,1]$,
\begin{equation}
u_t+uu_x-\nu u_{xx}=0,
\qquad
\nu=\frac{0.01}{\pi},
\end{equation}
where subscripts denote partial derivatives and $\nu$ is the viscosity, with $u(x,0)=-\sin(\pi x)$ and $u(-1,t)=u(1,t)=0$. Final reporting uses the complete numerical reference grid used in our experiments.

\paragraph{Wave1D.}
For $(x,t)\in[0,1]^2$,
\begin{equation}
u_{tt}-4u_{xx}=0,
\end{equation}
where $u_{tt}$ and $u_{xx}$ denote the second derivatives with respect to time and space, respectively. The boundary values and initial
displacement are induced by the exact solution
\begin{equation}
\begin{aligned}
u^*(x,t)
={}&\sin(\pi x)\cos(2\pi t)\\
&+\tfrac12\sin(a\pi x)\cos(2a\pi t),
\qquad a=4,
\end{aligned}
\end{equation}
together with the initial velocity condition $u_t(x,0)=0$.

\paragraph{Heat2D.}
For $(x,y,t)\in[0,1]^2\times[0,5]$,
\begin{equation}
u_t-\kappa_xu_{xx}-\kappa_yu_{yy}=0,
\end{equation}
where $\kappa_x$ and $\kappa_y$ are the diffusivities along the $x$ and $y$ directions, with zero spatial boundary values and the exact initial condition induced by
\begin{equation}
u^*(x,y,t)=
\sin(k_xx)\sin(k_yy)
\exp[-(\kappa_xk_x^2+\kappa_yk_y^2)t],
\end{equation}
where
\begin{equation}
\kappa_x=(500\pi)^{-2},\quad
\kappa_y=\pi^{-2},\quad
(k_x,k_y)=(20\pi,\pi).
\end{equation}
Here $k_x$ and $k_y$ are the corresponding spatial wavenumbers. For Wave1D and Heat2D, final errors are evaluated on independently sampled
interior coordinates from the analytic solution.

\subsection{Reference Topologies of the Frozen Algorithms}
\label{app:frozen-algorithms}

The reporting experiments retrain fixed algorithms from scratch; they do not invoke the generator. Table~\ref{tab:frozen-algorithms} summarizes the principal representation topology, training mechanism, and fixed structural settings of each frozen algorithm.

\begin{table*}[t]
\centering
\small
\setlength{\tabcolsep}{3pt}
\begin{tabular}{
@{}
p{0.12\textwidth}
p{0.27\textwidth}
p{0.27\textwidth}
p{0.25\textwidth}
@{}
}
\toprule
Problem & Representation topology & Training mechanism & Fixed structural setting \\
\midrule
Poisson2D & Spectral-lifted gated residual network & Residual-lift Adam rewarming with periodic residual pressure and optional quasi-Newton refinement & Six frequency bands, width 100, five gated residual blocks \\
Burgers1D & Bilinear transport-interaction network & Adam preconditioning followed by quasi-Newton refinement & Width 100, five interaction blocks \\
Wave1D & Wave-phase lift with feature-wise modulated residual blocks & Causal-horizon curriculum & Six-frequency lift, width 100, seven modulated blocks \\
Heat2D & Time-modulated multiscale spatial representation & Localized residual-gradient targeting & Four spatial and three temporal bands, width 100 \\
\bottomrule
\end{tabular}
\caption{Frozen EvoPINN algorithms used for independent reporting retraining.}
\label{tab:frozen-algorithms}
\end{table*}

\subsection{SLRC-PINN Implementation Notes}
\label{app:SLRC-pinn}
\label{app:SLRC}

EvoPINN discovered SLRC-PINN during the Burgers1D search as an accepted representation change rather than through post-search manual design. The main paper gives the complete mathematical topology; here we record the implementation details needed to interpret that formulation. The input is $q=(x,t)\in\mathbb R^2$, and the basis bank contains six learned centers, log-scales, and coordinate-dependent amplitude, modulation, and scale projections. The raw coordinate and all six basis responses are concatenated before the gated residual backbone. Separate linear decoders produce the global prediction and local correction, while a learned scalar gate combines the basis responses to modulate only the corrective branch.

The final linear layer of the local decoder is initialized with zero weights and zero bias. Therefore $u_L(q)\equiv0$ and the initial network function is exactly the global branch, independent of the initial gate values. All basis centers, widths, projections, backbone parameters, decoders, and gate parameters are subsequently optimized jointly by the same physics-informed objective. No subdomain labels, interface losses, or manually specified localization masks are used.

\subsection{Controlled SLRC-PINN Comparison}

The controlled study compares SLRC-PINN with the parameter-matched global MLP, APINN, FBPINN, and HyResPINNs listed in the main paper. All methods use the same Burgers equation, Hammersley collocation, physics-informed mean-squared objective, 20,000 Adam--L-BFGS optimization steps, relative-$L_2$ scorer, and evaluation seeds $\{0,1,2,3,4\}$. Parameter counts are computed after construction and document the parameter-matched comparison; they do not alter the training budget. APINN, FBPINN, and HyResPINNs preserve their defining architectural structures under the shared PyTorch/DeepXDE runner rather than claiming exact reproduction of the numerical protocols in their original
studies.

SLRC-PINN is a mechanism-discovery case study and is distinct from the frozen Burgers1D algorithm used in the main benchmark table. The former isolates and tests the discovered global--local mechanism; the latter is selected strictly by the search objective for the principal comparison.

\subsection{TCROP: Telemetry-Conditioned Program Discovery}
\label{app:tcrop}

Beyond neural representation search, EvoPINN successfully synthesized an executable training policy for Poisson2D, termed the \textbf{Telemetry-Conditioned Resampling and Optimizer Policy (TCROP)}. This case study examines its formulation, structural mechanisms, and optimization dynamics. TCROP is a particularly revealing discovery because it shows that agentic evolution can produce a coordinated training program, rather than merely select an optimizer or tune an isolated scalar hyperparameter.

\subsubsection{Algorithmic Formulation and Policy Mapping}
TCROP maps the compressed diagnostic state $s_g$ of a parent run to a training schedule $\pi(s_g)=(B_{\mathrm{Adam}},B_{\mathrm{LBFGS}},r_{\mathrm{refresh}})$, where $B_{\mathrm{Adam}}$ and $B_{\mathrm{LBFGS}}$ are the step allocations for the two optimizers and $r_{\mathrm{refresh}}$ is the collocation-refresh period. The state is
\begin{equation}
\begin{aligned}
s_g=\bigl(&
z_{\mathrm{conv}},
z_{\mathrm{trend}},
z_{\mathrm{osc}},
\rho_{\mathrm{conv}},
\rho_{\mathrm{osc}},\\
&
s_{\mathrm{late}},
s_{\mathrm{recent}},
n_{\mathrm{log}},
d
\bigr).
\end{aligned}
\end{equation}
where $z_{\mathrm{conv}}$, $z_{\mathrm{trend}}$, and $z_{\mathrm{osc}}$ are the categorical convergence, trend, and oscillation labels produced by the deterministic diagnostic map in Appendix~\ref{app:diagnostics}; $\rho_{\mathrm{conv}}$ and $\rho_{\mathrm{osc}}$ are the continuous contraction and relative-oscillation statistics defined there; $n_{\mathrm{log}}$ is the number of valid logged checkpoints; and $d$ is the dominant-constraint label. The synthesized second-order readiness condition is
\begin{align}
\delta_{\mathrm{smooth}}
={}&\mathbb I[z_{\mathrm{osc}}=\text{low}]
\nonumber\\
&\lor
\mathbb I[\rho_{\mathrm{osc}}<0.03],
\\[0.2em]
\delta_{\mathrm{plateau}}
={}&\mathbb I[
|s_{\mathrm{late}}|<2.5\times10^{-4}
]
\nonumber\\
&\land
\mathbb I[
|s_{\mathrm{recent}}|<2.5\times10^{-4}
],
\\[0.2em]
\delta_{\mathrm{ready}}
={}&\delta_{\mathrm{smooth}}
\land
\bigl(
\delta_{\mathrm{conv}}
\lor
\delta_{\mathrm{plateau}}
\bigr)
\nonumber\\
&\land
\mathbb I[n_{\mathrm{log}}\ge10].
\end{align}
where $\lor$ and $\land$ denote logical OR and AND; $s_{\mathrm{late}}$ and $s_{\mathrm{recent}}$ are least-squares loss slopes
over the fixed late and most-recent checkpoint windows, respectively; and $\delta_{\mathrm{smooth}}$, $\delta_{\mathrm{plateau}}$, and
$\delta_{\mathrm{ready}}$ are binary indicators of smoothness, plateauing, and second-order readiness. Finally, $\delta_{\mathrm{conv}}=\mathbb I[z_{\mathrm{conv}}=\text{high}]$ indicates a well-converged residual trajectory.

For the reported Poisson2D trajectory, the logged diagnostic state satisfies $\delta_{\mathrm{ready}}=1$. TCROP therefore adjusts the standard training allocation. Under a total nominal resource envelope of $B=20{,}000$ steps, it allocates:
\begin{equation}
B_{\mathrm{Adam}} = 17{,}200, \qquad B_{\mathrm{LBFGS}} = 2{,}800,
\end{equation}
paired with an Adam learning rate of $5\times 10^{-4}$ and an active global collocation refresh gate enabled at period $r_{\mathrm{refresh}}=1{,}000$.

\subsubsection{Mechanistic Advantage: Programmatic Synergy of Policy and Resampling}
To interpret TCROP's superior accuracy, we examine the two mechanisms that the generated program coordinates. Rather than relying on a rigid, static optimizer handoff, TCROP combines:

\begin{enumerate}
    \item \textbf{Telemetry-Driven Budget Allocation:} Under the observed readiness state, TCROP extends the initial Adam phase to $B_{\mathrm{Adam}}=17{,}200$ steps before assigning $B_{\mathrm{LBFGS}}=2{,}800$ steps to quasi-Newton refinement. The resulting combined policy substantially outperforms the conventional equal-split baseline.
    \item \textbf{Synergistic Global Collocation Renewal:} During Adam, TCROP activates global resampling every $1{,}000$ steps, producing 17 point renewals across the interior and boundary domains. These renewals repeatedly expose the optimizer to fresh physical constraints and are designed to reduce early over-specialization to a fixed collocation set before L-BFGS.
\end{enumerate}

This programmatic synergy highlights a key strength of EvoPINN: the framework does not merely tune scalar hyperparameters, but autonomously synthesizes holistic, execution-grounded training policies from observed convergence behavior.

\subsubsection{Two-Phase Optimization Dynamics: Stochastic Preconditioning and Second-Order Refinement}
We empirically validate the discovered TCROP policy against standard optimization baselines across five paired random seeds under identical initializations and resource envelopes (Table~\ref{tab:tcrop-baselines}). The discovered TCROP policy achieves a \textbf{27.5\% reduction in geometric-mean relative $L_2$ error} compared to the conventional $10\text{k}/10\text{k}$ handoff, while simultaneously lowering seed variability ($\mathrm{CV} = 19.0\%$).

\begin{table}[h!]
\centering
\footnotesize
\setlength{\tabcolsep}{2.5pt}
\begin{tabular}{@{}lrrrr@{}}
\toprule
Protocol & GM Rel-$L_2$ & Median & CV & Time \\
\midrule
20k Adam & 2.1104e-1 & 1.9747e-1 & 25.2\% & 19.1 \\
Fixed 10k/10k & 6.9808e-3 & 6.3477e-3 & 29.6\% & 32.9 \\
TCROP & \textbf{5.0588e-3} & \textbf{5.2331e-3} & \textbf{19.0\%} & 34.4 \\
\bottomrule
\end{tabular}
\caption{Controlled Poisson2D comparison over five paired seeds. GM denotes geometric mean, and time is measured in minutes. TCROP dynamically discovers and executes the $17.2\text{k}/2.8\text{k}$ optimizer allocation paired with gated global collocation renewals.}
\label{tab:tcrop-baselines}
\end{table}

\paragraph{Optimization-Trajectory Analysis.}
The optimizer-transition statistics provide a direct view of how the two
protocols enter L-BFGS:
\begin{itemize}
    \item \textbf{Substantially Lower Entry Loss:} At step $17{,}200$, the geometric-mean training loss under TCROP reaches $6.90 \times 10^{-2}$. In contrast, the conventional handoff at step $10{,}000$ leaves an entry loss of $2.81 \times 10^{-1}$—which is \textbf{$4.1\times$ higher}.
    \item \textbf{Precondition-and-Polish Effect:} The extended Adam phase and periodic renewals deliver a markedly lower-loss state to L-BFGS. Together with the lower final error, this supports the interpretation that TCROP creates a more favorable entry point for second-order refinement.
\end{itemize}

\paragraph{Execution-Fidelity Control.}
A stateless control executing the exact discovered $17.2\text{k}/2.8\text{k}$ schedule and refresh actions reproduces TCROP's accuracy ($5.06\times10^{-3}$). This control verifies that the gain is carried by the executable program emitted by the search rather than by hidden
outer-loop state at reporting time. It is not intended to isolate the individual causal contribution of the budget split and resampling; the scientific claim here concerns the effectiveness and reproducibility of the discovered joint policy.

\section{Implementation and Evaluation Protocols}
\label{app:reproducibility}

\subsection{Unified Budget and Evaluation Protocol}
\label{app:budget}

The generation-specific resource envelope is
\begin{equation}
B_g=
\left(
N_{r,g}^{(0)},N_{b,g}^{(0)},N_{i,g}^{(0)},K_{\mathrm{train},g},
\mathcal O_g,\mathcal Q_g
\right),
\end{equation}
where the sample counts specify the initial counts of interior, boundary, and initial constraints, $K_{\mathrm{train},g}$ is the total optimization-step limit,
$\mathcal O_g$ is the permitted optimizer family, and $\mathcal Q_g$ contains the configured adaptive-sampling controls. All candidates within a generation share the same $B_g$, which may be adjusted between generations. A candidate may reorganize optimization phases, losses, and sampling using these controls, but it may not alter the PDE, physical constraints, reference points, score definition, or total step budget. Matching the envelope aligns the principal training resources; it does not imply identical wall-clock time or floating-point operation count.

Search-time selection and final reporting use distinct reference point sets. The outer loop can access only the search set used to score candidates. After selection, the algorithm is frozen, retrained from scratch with previously unused reporting seeds, and evaluated on independently prepared reporting points. No reporting coordinate is used for candidate selection. This separation prevents the main reported results from reusing the point set that drives outer-loop evolution.

\begin{table}[t]
\centering
\footnotesize
\setlength{\tabcolsep}{4pt}
\begin{tabular}{@{}lrr@{}}
\toprule
Source
& \shortstack{Execution\\time (h)}
& \shortstack{GPU training\\time (h)} \\
\midrule
Poisson2D & 48.90 & 41.07 \\
Burgers1D & 16.72 & 14.70 \\
Wave1D & 70.88 & 51.78 \\
Heat2D & 24.51 & 21.92 \\
\bottomrule
\end{tabular}
\caption{Approximate cost recorded for the designated source search of each
benchmark.}
\label{tab:search-runtime}
\end{table}

\begin{table*}[t]
\centering
\small
\begin{tabular}{p{0.25\textwidth}p{0.25\textwidth}p{0.40\textwidth}}
\toprule
Protocol component & Setting & Role \\
\midrule
Nominal full-budget setting & $N_r^{(0)}=8192$, $N_b^{(0)}=2048$, $N_i^{(0)}=2048$; 20,000 optimization steps & Common reporting configuration and nominal full-budget search setting; unused condition counts are omitted for steady problems \\
Point construction & Deterministic Hammersley sampling & Common physics-constraint sampling rule \\
Configured adaptive-sampling controls & Nominal period 1000; update size 100; candidate pool 3000 & Program-dependent controls for refresh, replacement, or anchor addition \\
Proposal batch and search horizon & Eight proposals per generation; seven generations & Common outer-loop opportunity budget \\
Program generation & gpt-5.4-mini; temperature $0.7$ & Independent generation of each proposal \\
Retry and repair limit & At most three API retries or validation-repair calls & Bounds generation recovery and validation-guided self-repair \\
Candidate time limit & 7200 seconds per isolated full evaluation & Failed, timed-out, and non-finite runs are ineligible for selection \\
Scheduler history and exploration & Eight recent observations per module; $\beta=0.35$ & Recent-window UCB in Equation~\eqref{eq:module-ucb} \\
Scheduler reward weights & $(\omega_a,\omega_\delta,\omega_v,\omega_n)=(0.55,0.20,0.15,0.10)$ & Acceptance, normalized improvement, executability, and novelty terms described in Appendix~\ref{app:scheduler} \\
Seed PINN & Five hidden layers of width 100, tanh, Glorot-normal initialization; 10,000 Adam steps followed by 10,000 quasi-Newton steps & Shared non-evolved reference algorithm \\
Reporting protocol & Seeds 2027--2031 for the seed PINN, selected expert, and frozen EvoPINN & Independent retraining and aggregate reporting \\
\bottomrule
\end{tabular}
\caption{Implementation hyperparameters collected in one place. These values specify the reported protocol but are not presented as universal properties of EvoPINN.}
\label{tab:implementation-hyperparameters}
\end{table*}

\subsection{Search, Retraining, and Expert Controls}

Three independent searches are run for each PDE. The principal frozen algorithm is the best archived algorithm from the designated source search with seed 1234, chosen by its lowest full-evaluation $S_{\mathrm{search}}$ before any reporting retraining. The remaining searches provide robustness evidence and do not enlarge the selection pool for the main comparison. The seed PINN, selected expert, and frozen EvoPINN are then retrained from scratch with the same five reporting seeds. Expert configurations are selected with seed 1234 during a separate screening stage; screening runs are excluded from reported expert statistics. Ablation and neighboring-instance experiments follow the budgets stated in the main paper and do not replace the principal reporting protocol. In the neighboring-instance experiments, the discovered representation and training-program modules (and the selected expert identities) are kept fixed, while network parameters are reinitialized and trained from scratch at each shifted parameter value; no additional algorithm search is performed.

Candidate evaluations run in isolated processes and must complete within the configured time limit with a finite score. Failed or timed-out candidates cannot enter selection. Random states for the numerical and learning frameworks are reset before each evaluation. This controls the intended random initial conditions but does not imply bitwise-identical accelerator execution.

\subsection{Prompt and Interaction Protocol}

Each proposal prompt supplies the PDE, parent algorithm, selected module, mechanism focus, resource envelope, diagnostic state, and summarized search history. It requests a complete executable replacement while prohibiting changes to the benchmark, reference data, scoring rule, and unchanged module. Repair prompts add validation evidence from the failed proposal but never reveal reference-solution values or reporting results. Initial and repaired proposals pass the same structural, interface, resource, and execution gates.

\subsection{Hardware and Computational Cost}

The experiments were executed on NVIDIA GeForce RTX 3090 GPUs with 24\,GB of memory using DeepXDE 1.15.0 and PyTorch 2.6.0 under a common Python, NumPy, CUDA, and cuDNN environment. The approximate execution costs of the four designated source-search runs are reported in
Table~\ref{tab:search-runtime}. These totals include both failed and successful candidate attempts and therefore represent the observed search cost under our experimental environment rather than a hardware-independent complexity measure.

\end{document}